%% file: main.tex
\documentclass[10pt,twocolumn,letterpaper]{article}

\usepackage[pagenumbers]{cvpr} 

\input{preamble}

\definecolor{cvprblue}{rgb}{0.21,0.49,0.74}
\usepackage[pagebackref,breaklinks,colorlinks,allcolors=cvprblue]{hyperref}

\usepackage[ruled,vlined,linesnumbered]{algorithm2e}
\usepackage{algpseudocode}
\usepackage{multirow}
\usepackage{amsmath}
\usepackage{wrapfig}
\usepackage{animate}
\usepackage{graphicx}
\usepackage{tikz}
\usepackage[accsupp]{axessibility}
\usepackage{bm} 

\def\paperID{551} 
\def\confName{CVPR}
\def\confYear{2025}

\title{Visual Prompting for One-shot Controllable Video Editing without Inversion}

\author{
    Zhengbo Zhang\textsuperscript{\rm 1}\thanks{Equal contribution.} \quad
    Yuxi Zhou\textsuperscript{\rm 2}\footnotemark[1] \quad
    Duo Peng\textsuperscript{\rm 1} \quad
    Joo-Hwee Lim\textsuperscript{\rm 3} \quad \\
    Zhigang Tu\textsuperscript{\rm 2}\thanks{Corresponding author: tuzhigang@whu.edu.cn} \quad
    De Wen Soh\textsuperscript{\rm 1} \quad
    Lin Geng Foo\textsuperscript{\rm 1} \quad
    \\
    \textsuperscript{\rm 1}Singapore University of Technology and Design \quad
    \textsuperscript{\rm 2}Wuhan University \quad \\
     \textsuperscript{\rm  3} Institute for Infocomm Research, Agency for Science, Technology and Research, Singapore  
    \quad \\
    \url{https://vp4video-editing.github.io/}
}

\begin{document}
\maketitle

\input{real_sec/0_abstract}
\input{real_sec/1_intro}

\input{real_sec/2_related_work}

\input{real_sec/3_preliminaries}

\input{real_sec/4_proposed_method}
\input{real_sec/5_experiment}

\input{real_sec/6_conclusion}

{
    \small
    \bibliographystyle{ieeenat_fullname}
    \bibliography{main}
}

\end{document}

%% file: preamble.tex
\definecolor{yellow}{rgb}{1, 1, 0.7}
\definecolor{orange}{rgb}{1, 0.85, 0.7}
\definecolor{tablered}{rgb}{1, 0.7, 0.7}

%% file: real_sec/0_abstract.tex
\begin{abstract}
One-shot controllable video editing (OCVE) is an important yet challenging task, aiming to propagate user edits that are made -- using any image editing tool -- on the first frame of a video to all subsequent frames, while ensuring content consistency between edited frames and source frames. To achieve this, prior methods employ DDIM inversion to transform source frames into latent noise, which is then fed into a pre-trained diffusion model, conditioned on the user-edited first frame, to generate the edited video. However, the DDIM inversion process accumulates errors, which hinder the latent noise from accurately reconstructing the source frames, ultimately compromising content consistency in the generated edited frames. To overcome it, our method eliminates the need for DDIM inversion by performing OCVE through a novel perspective based on visual prompting. Furthermore, inspired by consistency models that can perform multi-step consistency sampling to generate a sequence of content-consistent images, we propose a content consistency sampling (CCS) to ensure content consistency between the generated edited frames and the source frames.
Moreover, we introduce a temporal-content consistency sampling (TCS)  based on Stein Variational Gradient Descent to ensure temporal consistency across the edited frames. Extensive experiments validate the effectiveness of our approach.
\end{abstract}

%% file: real_sec/1_intro.tex
\section{Introduction}
\label{sec:intro}

Video production plays a crucial role in creating compelling visuals for films, short videos, and various other media formats, its significance has rapidly increased amidst the ever-growing demand for high-quality video content.
For instance, high-quality video production is often highly important for bloggers to create entertaining video vlogs on social platforms~\cite{wibowo2021cinematic} or for filmmakers to generate captivating virtual scenes in films~\cite{wu2017analyzing}.
Yet, in many cases, both generated and real-world video content may fall short of meeting specific user requirements. As a result, there has been a significant increase in demand for convenient \emph{video editing} tools that allow users to modify videos according to customized instructions.

Recently, diffusion-based video editing methods~\cite{liew2023magicedit,xing2024simda,lee2023shape,ceylan2023pix2video,qi2023fatezero} have gained considerable attention. 
These approaches have demonstrated strong performance across various video editing tasks, such as visual style transfer~\cite{ceylan2023pix2video} and character or object modification~\cite{qi2023fatezero}. Additionally, they are highly user-friendly, often requiring only minor adjustments to the textual descriptions of the video content.
Yet, although video editing via modifying textual descriptions offers convenience, this approach is a double-edged sword, as it typically facilitates global changes which often restricts the \emph{controllability} of the editing. 
For instance, consider a scenario where social media users aim to make precise modifications to specific objects, structures or layouts in the video. Achieving such fine-grained control is challenging when relying solely on text-based descriptions. Therefore, controllable video editing methods are required to achieve the desired outcomes in such cases.

Nevertheless, achieving controllable video editing beyond text-based instructions is challenging, as it requires both high versatility and precision, including capabilities like accurately repositioning objects, erasing or adding specific elements, \etc. 
To enable accurate and efficient controllable video editing, \textit{one-shot controllable video editing} (OCVE) approaches~\cite{gu2024videoswap,ku2024anyv2v,fan2024videoshop} have been introduced. 
These approaches allow users to apply desired edits to the first frame of the source video using any off-the-shelf image editing tools (\eg, Photoshop, Paint, image editing diffusion models), and these approaches then propagate the edits to the remaining video frames. 
These OCVE methods commonly utilize \textit{DDIM inversion}~\cite{song2020denoising}, which is a recursive process that converts the source video into latent noise, serving as latent representations that enable the diffusion models to reconstruct the source video.
Then, the inverted latents are combined with editing guidance (which indicates the target parts or objects to edit), and processed through the diffusion's sampling process to generate the edited video, thereby achieving controllable video editing. 
By leveraging the DDIM inversion in this manner, such  OCVE approaches can achieve efficient editing while preserving the information of the source video.

Although the aforementioned DDIM inversion-based OCVE methods have made commendable progress, they still encounter the following two challenges. 
Firstly, the DDIM inversion introduces approximation errors at each timestep~\cite{song2020denoising}, and the accumulation of these errors often degrades the quality of the reconstructed video, which, in turn, diminishes the \textit{content consistency} of the edited video and weakens the editing capabilities of DDIM inversion-based methods~\cite{xu2024task}. 
Secondly, after obtaining the inverted latents by DDIM inversion, when using image diffusion models~\cite{rombach2022high} to generate edited video frames, the lack of strong temporal priors can lead to edited videos having poor \emph{temporal consistency}~\cite{yang2023rerender,chai2023stablevideo}. 
To tackle this, some methods~\cite{fan2024videoshop,ku2024anyv2v} utilize video diffusion models~\cite{blattmann2023stable,zhang2023i2vgen} to provide temporal priors. However, the quality of these temporal priors is often insufficient, since the available open-sourced video datasets~\cite{xue2022advancing,bain2021frozen,miech2019howto100m} used to train the video diffusion models tend to be of lower quality and size compared to the proprietary datasets used to train high-quality, closed-sourced models like Sora~\cite{sora} and Kling~\cite{Kling}. 
Besides, video diffusion models are computationally demanding, resulting in the approaches based on these models being highly time-consuming.

\input{image/tex/new_perspective}
Therefore, in this paper, we eschew the DDIM inversion process which can potentially introduce errors, and instead approach OCVE from a novel perspective, by treating it as a \textit{visual prompting task}~\cite{bar2022visual}. 
Our insight is that, both OCVE and visual prompting share the goal of propagating certain modifications across images (See~\cref{fig:new perspective}).
From this perspective, to tackle OCVE, we can consider the first source frame and  the first edited frame (modified using any image editing tool), as the visual prompting example, with each remaining frame of the source video as a query.
The visual prompting example and query are subsequently input into a diffusion model to obtain the edited frame, ensuring that the image pair -- comprising the query and the output edited frame -- remains consistent with the provided example. 
At the same time, we employ a pre-trained inpainting image diffusion model~\cite{inpaint2022} to achieve it, leveraging its strong visual reasoning ability.
Our method \emph{bypasses the inaccurate DDIM inversion},  because the query frame that requires editing is directly input into the diffusion model in the encoded feature representation using the diffusion model's image encoder, rather than being inverted into latent noise by the DDIM inversion.

Additionally, to facilitate the edited frames to better maintain \emph{content consistency} with the source frames, we draw inspiration from consistency models~\cite{song2023consistency} which employ multi-step consistency sampling to generate a sequence of content-consistent images, and we introduce a content consistency sampling (CCS) method. 
Furthermore, we develop a temporal-content consistency sampling (TCS) method based on Stein Variational Gradient Descent (SVGD)~\cite{liu2016stein} to ensure that the generated frames also exhibit good \emph{temporal consistency}.
Notably, compared to methods~\cite{ku2024anyv2v,fan2024videoshop} that depend on video diffusion models to provide temporal priors for preserving the temporal consistency of edited frames, our TCS is much faster while also providing quality improvements. Experimental results indicate that our method achieves strong performance in OCVE.

%% file: image/tex/new_perspective.tex
\begin{figure}[t!]
  \centering
\includegraphics[width=0.48\textwidth]{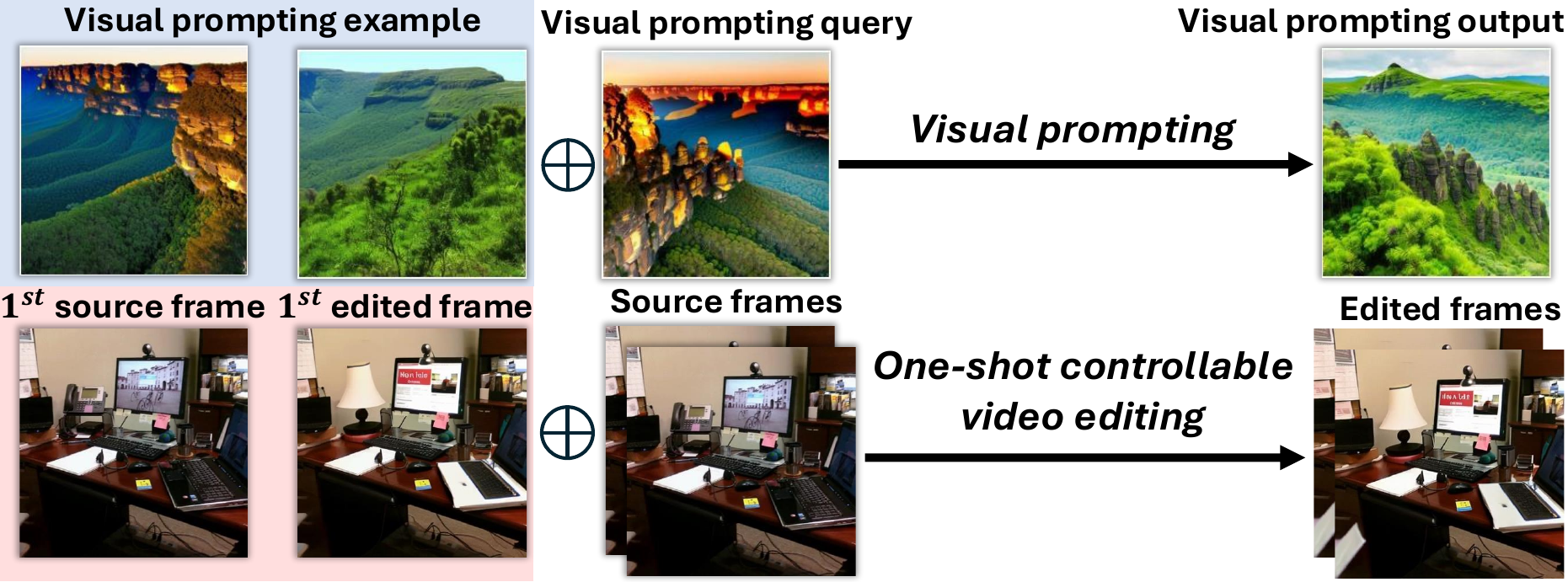}
  \caption{Visual prompting and one-shot controllable video editing share the goal of propagating certain modifications across images. In visual prompting, modifications made in the example (\eg, changing the color of the mountain from golden to green) are transferred to the query, whereas in one-shot controllable video editing, modifications applied to the first edited frame are propagated to the subsequent source frames. 
  }
  \label{fig:new perspective}
\end{figure}

%% file: real_sec/2_related_work.tex
\section{Related work}
\label{sec:related work}

\noindent \textbf{Video editing}~\cite{bar2022text2live,kasten2021layered} is a challenging task that aims to modify the content of a given video according to the user's intentions. 
Recently, inspired by the extensive knowledge contained in pre-trained diffusion models (e.g., Stable Diffusion~\cite{rombach2022high}), researchers have widely explored leveraging diffusion models to facilitate video editing. 
Yet, most existing diffusion-based video editing approaches~\cite{yatim2024space,qi2023fatezero,kahatapitiya2024object,kara2024rave,shin2024edit,ren2024customize,zuo2023cut,jeong2023ground,couairon2023videdit,geyer2023tokenflow,yang2023rerender,zhu2024zero,wu2023tune,liu2024video,cong2023flatten,ouyang2024codef,zhang2024fastvideoedit,kim2023collaborative} are predominantly text-driven, where videos are modified based mostly on textual instructions for tasks such as style editing~\cite{ceylan2023pix2video}, object editing~\cite{qi2023fatezero}, motion transfer~\cite{yatim2024space}, texture editing~\cite{couairon2023videdit}, \etc. 
Since text-based video editing methods often face challenges in achieving fine-grained controllability, controllable video editing~\cite{yoon2024raccoon,deng2023dragvideo,ma2023magicstick} methodologies have recently drawn significant attention. 
In particular, such controllable video editing approaches can be categorized according to their training data requirements: methods requiring abundant training data~\cite{yoon2024raccoon,mou2024revideo}, few-shot methods~\cite{deng2023dragvideo}, and one-shot methods~\cite{ma2023magicstick,gu2024videoswap,ku2024anyv2v,fan2024videoshop}.

In this paper, we focus on the \textit{controllable one-shot setting} (OCVE), which aims to edit the source video given only one edited frame (which shows the exact desired edits by the user).
Existing one-shot methods~\cite{ma2023magicstick,ku2024anyv2v} often rely on DDIM inversion~\cite{song2020denoising} to obtain the inverted latent for video editing, which facilitates video editing by enabling the diffusion model to first approximately reconstruct the source video. 
However, the inherent approximation errors in DDIM inversion limits the editing capabilities, compromising quality and content consistency in the edited video~\cite{xu2024task}.
Different from previous works, we eschew DDIM inversion by approaching OCVE from a novel visual prompting perspective. Furthermore, we propose CCS, which is based on the multi-step consistency sampling of consistency models to improve content consistency, as well as TCS based on Stein Variational Gradient Descent~\cite{liu2016stein} to improve temporal consistency. Overall, this results in significantly improved quality for OCVE.

\noindent \textbf{Diffusion models}~\cite{song2020denoising,ho2020denoising,song2020score,foo2023ai} have demonstrated remarkable success in image generation, by learning to progressively refine samples from a tractable noise distribution towards the target data distribution. 
Because of their impressive performance, researchers~\cite{guo2025motionlab,hertz2022prompt,gu2024analogist,inpaint2022,kahatapitiya2024object,gong2023diffpose,wang2023modelscope,xu2024task,foo2024action,yang2023rerender,chai2023stablevideo,guo2023animatediff,guo2025tstmotion,foo2025avatar} have applied diffusion models to a range of tasks, including image inpainting~\cite{inpaint2022}, image editing~\cite{ju2024pnp}, pose estimation~\cite{foo2023distribution}, video editing~\cite{kahatapitiya2024object}, and video generation~\cite{wang2023modelscope}, leading to significant advancements in these areas. 
In this paper, unlike previous works in video editing~\cite{kahatapitiya2024object,ku2024anyv2v}, we leverage the strong visual reasoning ability of an image inpainting diffusion model~\cite{inpaint2022} to perform OCVE through visual prompting. \
Besides, we modify the sampling process of the inpainting diffusion model to emulate the multi-step consistency sampling of consistency models, ensuring that the generated edited frames maintain content consistency with the source frames.

\noindent \textbf{Consistency models}~\cite{song2023consistency,luo2023latent} are a new class of generative models designed for fast sampling, which allows for efficient one-step generation. 
A key characteristic of consistency models is self-consistency, which ensures that samples generated along a trajectory can be directly mapped back to their initial state. 
In addition to their one-step generation capabilities, consistency models also support multi-step consistency sampling~\cite{song2023consistency,luo2023latent}, which provides a trade-off between computational efficiency and sample quality, enabling the generation of a sequence of content-consistent images. 
In this work, for the first time, we modify the sampling equations of a pre-trained inpainting diffusion model to enable the multi-step consistency sampling without requiring additional training.

%% file: real_sec/3_preliminaries.tex
\input{image/tex/pipeline}

\section{Preliminaries}
\label{subsec:Preliminaries}
OCVE is a challenging task, requiring the preservation of both the realism and consistency of the edited video with the source video, while editing the video according to the user's intentions.
Given that diffusion models~\cite{song2020denoising,rombach2022high} -- trained on large datasets of real images and videos -- exhibit a strong ability to generate highly realistic visuals, researchers~\cite{ku2024anyv2v,fan2024videoshop} often leverage their strong capabilities to address the challenging OCVE task. 
Notably, to enhance efficiency, the source video is processed in its latent representations, where the latent diffusion model is used.
Below, we detail the process of mapping the source video into the latent space (\ie, \textit{inverting the video into a noise latent}) to obtain a source latent that aids the diffusion model in reconstructing the source video. We first provide an overview of how latent diffusion models typically generate samples, followed by how the inversion is often done.

\noindent
\textbf{Latent diffusion models} produce target samples ($z_0$) via a progressive latent denoising process consisting of $T$ recursive steps. Specifically, starting from $t = T$, the $t$-th denoising step denoises a latent at the $t$-th step ($z_{t}$) into a latent at the $(t-1)$-th step $(z_{t-1})$ as follows~\cite{ho2020denoising}:
\begin{equation}
\begin{split}
\label{eq:diffusion sampling}
z_{t-1}=\sqrt{\alpha_{t-1}} \cdot \underbrace{(z_t-\sqrt{1-\alpha_t} \cdot\epsilon_\theta(z_t,t)) /\sqrt{\alpha_t})}_{\text { predicted } \hat{z}_0 }\\+ \underbrace{\sqrt{1-\alpha_{t-1}-\sigma_t^2}  \cdot\epsilon_\theta(z_t,t)}_{\text {adjustment along } z_t }+\underbrace{\sigma_t \cdot \epsilon}_{\text {random noise}}, 
\end{split}
\end{equation}
where $\epsilon_\theta(z_t, t)$ is a noise prediction network parameterized by $\theta$ (often adopting a U-Net~\cite{ronneberger2015u} autoencoder),  $\alpha_{1: T}\in \left(0, 1\right)$ is a decreasing sequence of coefficients, $\sigma_{1:T}$ is a noise schedule,  and $\epsilon \sim \mathcal{N}(0, I)$ is standard Gaussian noise independent of $z_t$.  
By applying Eq.~\ref{eq:diffusion sampling} recursively for $T$ steps, we eventually obtain $z_0$ as an output denoised latent.

Due to the iterative nature of the denoising diffusion process (from $z_T$ to $z_0$) explained above, the straightforward inversion approach to map the source video into the latent space, \ie, compute $z_T$ from $z_0$, would also be an iterative process, where the $t$-th step is given by:
\setlength{\abovedisplayskip}{4pt}
\setlength{\belowdisplayskip}{3pt}
\begin{equation}
\begin{aligned}
\label{eq:ddim inversion}
z_{t}=&(\sqrt{\alpha_{t}}\cdot z_{t-1}-\sqrt{\alpha_{t}}\sqrt{1-\alpha_{t-1}}\cdot\epsilon_\theta(z_t,t)) / \sqrt{\alpha_{t-1}}\\ + &\sqrt{1-\alpha_{t}}\cdot\epsilon_\theta(z_t,t)
\end{aligned}
\end{equation}
Here, the noise schedule $\sigma_t$ is set to $0$, eliminating the ``random noise'' item in~\cref{eq:diffusion sampling} and  transforming~\cref{eq:diffusion sampling} into a deterministic forward process, which facilitates the inversion process.
Yet, as evident from~\cref{eq:ddim inversion}, directly performing the inversion process is impractical because the noise prediction network $\epsilon_{\theta}(\cdot,\cdot)$ requires the desired $z_t$ as input.

\noindent
\textbf{DDIM inversion}~\cite{song2020denoising}  has been proposed to solve this inversion issue, by assuming that the ordinary differential equation process can be reversed in the limit of  infinitesimally small timesteps. 
Concretely, in the DDIM inversion process, $ \epsilon_\theta(z_t,t) $ is replaced with $ \epsilon_\theta(z_{t-1},t) $ for the noise prediction in~\cref{eq:ddim inversion}, which makes it tractable. 
However, although approximating $ \epsilon_\theta(z_t,t) $ with $ \epsilon_\theta(z_{t-1},t) $ achieves inversion to some extent, this approximation introduces errors at each timestep. Such cumulative errors degrade the reconstruction quality of latents $ z_t, \dots, z_1 $, which can ultimately diminish the performance of video editing methods~\cite{xu2024task}.

%% file: image/tex/pipeline.tex
\begin{figure*}[t!]
  \centering
\includegraphics[width=1\textwidth]{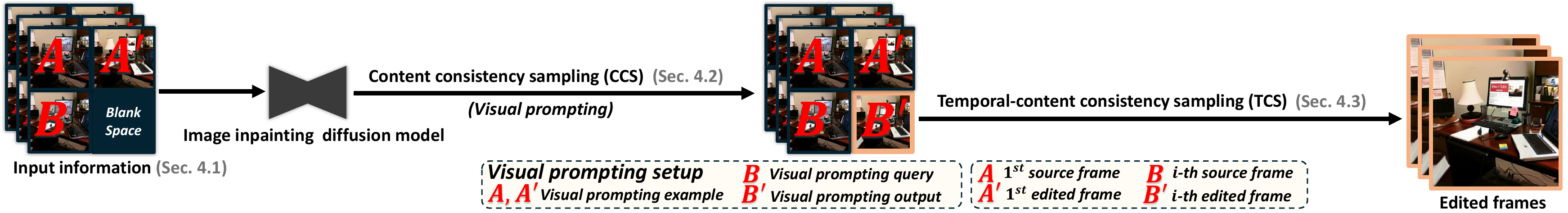} 
  \caption{
  The overall pipeline of our method. \textbf{First}, to enable the image inpainting diffusion model to perform OCVE through visual prompting, we organize the example (\boldsymbol{$A$} and \boldsymbol{$A'$}), query (\boldsymbol{$B$}), and output (\boldsymbol{$B'$}) from the visual prompting setup into a 2$\times$2 square grid, which serves as the input information (\cref{subsec:Performing our task via visual prompting}) to the inpainting diffusion model. \textbf{Next}, we modify the sampling process of the inpainting diffusion model, and design a content consistency sampling (\cref{subsec:consistency sampling}),  to generate \(\boldsymbol{B'}\) using the multi-step consistency sampling of the consistency models~\cite{song2023consistency}. \textbf{Finally}, based on the generated \(\boldsymbol{B'}\), we apply Temporal-content Consistency Sampling (\cref{subsec:SVGD})  with Stein Variational Gradient Descent~\cite{liu2016stein} to adjust the source frames, enhancing their temporal consistency and yielding the final edited frames in our framework. 
  }
  \label{fig:pipeline}
\end{figure*}

%% file: real_sec/4_proposed_method.tex
\section{Method}
\label{sec:proposed method}
In this work, to avoid relying upon DDIM inversion~\cite{song2020denoising} which may introduce errors, for the first time, we approach OCVE from a visual prompting perspective~(\cref{subsec:Performing our task via visual prompting}).
See ~\cref{fig:pipeline} for a summary of our full pipeline.
In our method, edited frames are generated as the visual prompting output by using the source frame as the visual prompting query to prompt an image inpainting diffusion model with the visual prompting example (the first source frame and the first edited frame).
To ensure the produced edited frames preserve content consistency with the source frames, we modify the sampling process of the inpainting diffusion model and propose a Content Consistency Sampling (CCS)  (\cref{subsec:consistency sampling}), leveraging the multi-step consistency sampling property of consistency models~\cite{song2023consistency} which can generate a sequence of content-consistent images. 
Finally, to make the output edited frame of our method maintain temporal consistency, we perform a Temporal-content Consistency Sampling (TCS)  (\cref{subsec:SVGD}) based on Stein
Variational Gradient Descent (SVGD)~\cite{liu2016stein}. In TCS, the edited frames produced by CCS are adjusted to align with the source frames, which helps preserve the temporal consistency.

\subsection{New perspective on OCVE}
\label{subsec:Performing our task via visual prompting}

In this paper, we approach OCVE from a fresh perspective, adopting a visual prompting approach and eschewing the DDIM inversion step~\cite{song2020denoising}.
Our key insight is that: both OCVE and visual prompting can both be understood as tasks \emph{focusing on the propagation of certain modifications} (See \cref{fig:new perspective}). 
In OCVE, the goal is to propagate user edits from the first frame to subsequent frames of the source video. In visual prompting, the aim is to propagate the modifications observed in an input-output image pair to the input query.
With this in mind, we observe that \textit{OCVE} can be re-casted as a type of a visual prompting task, where the given example consists of the edited and source versions of the first frame, and the query can be the subsequent frames of the source video. 
Notably, our approach does not require DDIM inversion because the source frames for editing are encoded as features, rather than as latent noise derived from the recursive DDIM inversion process.
In this subsection, we introduce each part of our visual prompting-based pipeline in more detail.

\noindent \textbf{Performing visual prompting with inpainting diffusion model.} 
To tackle the challenging OCVE task, we follow prior methods~\cite{ku2024anyv2v,fan2024videoshop} to leverage the extensive knowledge encoded in a pre-trained diffusion model.  The employed diffusion model is expected to have strong and robust visual reasoning capabilities, as we aim to perform visual prompting with the diffusion model, \ie, the diffusion model needs to infer how to propagate the frames of the source video based only on the provided pair of example frames (first edited frame +  first source frame). 
Here, we propose leveraging an image inpainting diffusion model for our work, as such models are well-suited for completing missing regions of an image while maintaining contextual consistency with the surrounding parts~\cite{inpaint2022}, demonstrating strong visual reasoning capabilities.
Yet, utilizing the diffusion model, originally designed for inpainting, to perform OCVE through visual prompting is not straightforward. 
To achieve it, we derive inspiration from~\cite{gu2024analogist}, and carefully \emph{tailor the inputs} to the inpainting diffusion model. In the following, we first introduce the inputs of the inpainting diffusion model, and then detail the modifications made to these inputs.

\noindent \textbf{Tailoring inputs of inpainting diffusion model.}  The image inpainting diffusion model is designed to achieve user-specified inpainting through a denoising process based on three input parameters.
  The parameters consist of: input information $G$ to be inpainted, mask information $M$ indicating the region in the input information $G$ that requires inpainting, and a guiding text prompt $p$ describing the desired inpainting result. To effectively harness the reasoning capability of the image inpainting diffusion model for OCVE through visual prompting, we design an inpainting strategy aligned with the model's original purpose, \ie, image inpainting, to generate the desired output (edited video frames). As illustrated in~\cref{fig:G(i)}, at the $i$-th source frame, the inpainting strategy organizes the input information $G(i)$ into four distinct regions: \textbf{(1)} the upper two regions hold the first frame before and after user editing, serving as the visual prompt example; \textbf{(2)} the bottom left region contains the $i$-th source frame, acting as the visual query; \textbf{(3)} the bottom right region is kept ``blank'', where we expect the diffusion model to generate the $i$-th edited frame via visual prompting based on the provided example and query. 
Additionally,~\cref{fig:G(i)} presents the corresponding mask information $M$ for the input information $G(i)$. In the mask information $M$, white regions denote the areas in the input information $G(i)$ where inpainting by the diffusion model is needed, while black regions indicate the corresponding areas in $G(i)$ that should remain unaltered.

\input{image/tex/G_i_}

We next describe the design of the text prompt $ p $, which is crucial for guiding the inpainting diffusion model toward generating the desired output, especially given the potential for multiple viable inpainting solutions.
In our task, intuitively, the text prompt $p$ may provide descriptions of the user's edits in the first frame. 
However, as described in~\cref{sec:intro}, accurately describing the user's editing in text is often challenging.
Given that vectors in the CLIP space can often effectively capture editing direction~\cite{parmar2023zero}, we represent the user's editing as the difference between the encoded features of the first edited frame $I^{e}(1)$ and first source frame $I^{s}(1)$ in the CLIP embedding space~\cite{radford2021learning}. 
Specifically, the user's editing, \ie, ``textual'' prompt $ p $  is calculated as:
\begin{equation}
\label{eq:prompt learning of p}
p = \lambda_1 \cdot  \{E_{CLIP}\big(I^{e}(1)\big) - E_{CLIP}\big(I^{s}(1)\big)\},
\end{equation}
where $ E_{CLIP}(\cdot) $ represents the image encoder of CLIP, and $\lambda_1$ is a hyper-parameter. 
Since our adopted inpainting diffusion model is built on the text encoder of CLIP, we can directly use $p$ as a ``text'' prompt to guide the diffusion model.

\subsection{Content consistency sampling (CCS)}

\label{subsec:consistency sampling}
In the previous subsection, we tackle OCVE from a visual prompting perspective, effectively bypassing the DDIM inversion step by generating the edited frames with an inpainting diffusion model. 
Ideally, the content of the generated edited frame should be based directly on the source frame to preserve content consistency between them. 
However, in our method, the content of the generated edited frame is based on the latent noise from the previous denoising timestep,  as our method builds upon the image inpainting diffusion model, where the edited frame is generated through a progressive denoising process~\cite{ho2020denoising} (see \cref{eq:diffusion sampling}).
  Moreover, since our method does not incorporate DDIM inversion, the initial noise in our progressive denoising process is not the noise obtained through DDIM inversion that approximately preserves the original image content. As a result, our method may face challenges in ensuring content consistency between the generated edited frame and the source frame.

To handle this, we get inspiration from consistency models, where their multi-step consistency sampling~\cite{song2023consistency,xu2023inversion} can generate content-consistent images by basing each timestep's  output image on the previous timestep's generated image. 
Based on this property, we propose a Content Consistency Sampling (CCS),
 which is a  multi-step consistency sampling built upon the progressive denoising sampling of the inpainting diffusion model. Specifically, to maintain content consistency between the CCS-generated edited frame and the source frame, we artificially configure CCS to generate the source frame in the first time step.
 A noise calibration mechanism is then applied to guide the sampling process, allowing the generated images to gradually transition from the source frame to the desired edited frame. 
 Note that CCS is a sampling approach, and thus does not require additional training.

 Below, we first introduce the special  sampling process of CCS, then we describe how our introduced sampling process is utilized to generate the desired edited frame with improved content consistency.

\noindent \textbf{Multi-step consistency sampling based on inpainting diffusion model.} 
In the multi-step consistency sampling of consistency models, the content of the generated image at each timestep is based on the output image from the previous denoising timestep, thereby generating a sequence of content-consistent images~\cite{luo2023latent}.
Here, we aim to utilize this property  to generate a sequence of content-consistent images, beginning with the source frame and progressively shifting the content  along the user's intended editing direction, ultimately generating a desired edited frame while preserving content-consistency with the source frame.
However, the sampling process in our inpainting diffusion model~(\cref{eq:diffusion sampling}) can struggle to generate a sequence of content-consistent images, due to its Markovian denoising nature~\cite{ho2020denoising}, which operates without reference to the source frame content.
Therefore, we seek to adapt the inpainting diffusion model’s denoising sampling into the multi-step consistency sampling. 
To this end, we first modify the inpainting diffusion model's sampling (\cref{eq:diffusion sampling}), so that it is no longer an iterative denoising Markov process. 
Then, we enable the modified sampling process to generate a latent of output image ($ \hat{z}_0 $) at each timestep, where the content of each generated latent is based on the content of the latent produced in the previous timestep, thereby yielding a sequence of content consistent images. 
These steps are detailed below.

Specifically, to eliminate  the Markovian denoising nature in the sampling of our inpainting diffusion model, we  remove the adjustment term (the second term) in~\cref{eq:diffusion sampling} by setting the noise parameter $ \sigma_t = \sqrt{1 - \alpha_{t-1}} $, 
resulting in the following  sampling:
\begin{equation}
\begin{split}
\label{eq:diffusion consistency sampling}
z_{t-1}=\sqrt{\alpha_{t-1}} \cdot \underbrace{(z_t-\sqrt{1-\alpha_t} \cdot\epsilon_\theta(z_t,t))/\sqrt{\alpha_t}}_{\textbf { predicted } \mathbf{\hat{z}_0} } \\ + \underbrace{\sqrt{1 - \alpha_{t-1}} \cdot \epsilon}_{\textbf {random noise}}, \epsilon \sim \mathcal{N}(0, I).
\end{split}
\end{equation}

Next, we consider the ``predicted $ \hat{z}_0 $'' as the output of the sampling process at each timestep, rather than the denoised latent (\eg, $z_{t-1}$ in~\cref{eq:diffusion consistency sampling}),  enabling the sampling process to output the latent of output image ($ \hat{z}_0 $) at each timestep following the multi-step consistency sampling.
Besides, in the multi-step consistency sampling, the generated latent ($ \hat{z}_0 $) across different timesteps should be consistent~\cite{song2023consistency}. However, the ``predicted $ \hat{z}_0 $'' at different timesteps are generally independent of each other, which may fail to ensure consistency across timesteps~\cite{li2023stimulating}. 
To solve it, we draw inspiration from ~\cite{xu2023inversion}, 
and introduce a consistency noise $ \epsilon^c $ to replace the parameterized noise $ \epsilon_\theta $ in the ``predicted $ \hat{z}_0 $'' term, ensuring that  $\hat{z}_0$ generated by the new term maintains consistency across different timesteps. Specifically, the new term is defined as:
\setlength{\abovedisplayskip}{4pt}
\setlength{\belowdisplayskip}{3pt}
\begin{equation}
\label{eq:assumed consistency model}
    \hat{f} (z_{t}, t, \epsilon^c(t)) = (z_{t} - \sqrt{1-\alpha_{t}} \cdot \epsilon^c (t)) / \sqrt{\alpha_{t}}, 
\end{equation}
and we have $\hat{z}_0^{(t)} = \hat{f} (z_{t}, t, \epsilon^c(t)).$ In fact, $ \hat{f} $ serves as a consistency model that can perform the multi-step consistency sampling. Hence, by substituting~\cref{eq:assumed consistency model} into~\cref{eq:diffusion consistency sampling}, we obtain the special multi-step consistency sampling, performed by iteratively executing the following process:
 \begin{equation}
\begin{split}
\label{eq:multi-step consistency model}
 & \hat{z}_{t-1}=\sqrt{\alpha_{t-1}} \cdot \hat{z}_0^{(t)} +  \sqrt{1 - \alpha_{t-1}} \cdot \epsilon, \epsilon \sim \mathcal{N}(0, I)  \\
 & \hat{z}_0^{(t-1)} = \hat{f} (\hat{z}_{t-1}, t-1, \epsilon^c (t-1)). \\
 \end{split}
\end{equation}
In this sampling process, the latent of the output frame generated at the current timestep (\eg, $\hat{z}_0^{(t-1)}$) is based on the latent of the output frame generated at the previous timestep ($\hat{z}_0^{(t)}$), thereby allowing the generation of a sequence of video frames that maintain content consistency.

\noindent \textbf{Generating desired edited frames.}
After deriving the special multi-step consistency sampling, we aim to to generate the desired edited frames by performing CCS. To this end, the sampling process of CCS starts by generating the source frame at the first timestep, ensuring that the generated edited frame maintains content consistency with the source frame. Next, a noise calibration mechanism is employed within the sampling process to gradually transform the generated frames from the source frame to the desired edited frame. We detail these steps below.

First, to ensure that CCS generates the source frame at the first timestep, it is necessary to determine the corresponding consistency noise $ \epsilon^c $. Since $ \hat{f} $ functions as a consistency model, we have $ z_0^s = \hat{f} (\hat{z}_{t}, t, \epsilon^c(t; z_0^s)) $, where $ z_0^s $ represents the latent of the source frame. Hence, from this equation, we can obtain the corresponding noise  $ \epsilon^c(t; z_0^s) = (\hat{z}_{t} - \sqrt{\alpha_t} \cdot z_0^s) / \sqrt{1 - \alpha_t}$.

Next, we aim to guide this multi-step consistency sampling to progress along the user's intended editing direction (\ie, guiding the content of the images generated by CCS to evolve in line with the user's intended edits), thereby generating the desired edited frame. 
Notably, this progression is reflected in the sampling process of the inpainting diffusion model used to generate the edited frame (\cref{subsec:Performing our task via visual prompting}). To be specific, in this sampling process, all regions initially receive the random Gaussian noise. However, the lower-left area of the input information $ G(i) $ is progressively denoised toward the source frame, while the lower-right area is denoised toward the desired edited frame. The denoising difference  between these two regions  at each timestep effectively captures the divergence between the source and edited frames, actually reflecting the user’s intended editing direction. 
Hence, we utilize this denoising difference  to guide CCS in generating images that evolve along the user's intended editing direction, ultimately generating the desired edited frame while maintaining content consistency with the source frame.

Specifically, the denoising difference $\Delta \epsilon_t$ is calculated as: $\Delta \epsilon_t = \epsilon_\theta(z_t(I^{e}),t) - \epsilon_\theta(z_t(I^{s}),t), $ where $ z_t(I^{e}) $ and $ z_t(I^{s}) $ denote the latent in the lower-right and lower-left regions of $ z_t $ at timestep $t$, respectively. We then add the denoising difference $\Delta \epsilon_t$ to the noise term $\epsilon_c$ of the consistency model $\hat{f}$ (\cref{eq:assumed consistency model}) in CCS at each timestep, as follows:
\begin{equation}
\label{eq:final:f_theta}
   \hat{f} (\hat{z}_t, t,  \epsilon^c (t;z_0^s)) = (\hat{z}_t - \sqrt{1-\alpha_t} ( \epsilon^c(t;z_0^s) + \lambda_2  \Delta \epsilon_t )) / \sqrt{\alpha_t}
\end{equation}
where $\lambda_2$ denotes a hyper-parameter.  Notably, CCS is performed on the lower-right subregion of the input information $G(i)$.

\subsection{Temporal-content consistency sampling (TCS)}
\label{subsec:SVGD}

Based on the CCS (\cref{subsec:consistency sampling}), we can perform OCVE without relying on DDIM inversion, and generate edited frames that maintain content consistency with the source frames.
However, CCS does not explicitly ensure the preservation of temporal consistency between source frames during its sampling process, which may lead to \emph{temporal inconsistency and unsmooth edited clips}. We address it by proposing a Temporal-consistent Consistency Sampling (TCS), which is performed following the completion of CCS.

To achieve this, we first explicitly model the temporal consistency of the source video by treating the video as a distribution, where each source frame is considered a sample drawn from this distribution. These source samples are constrained by their mutual relationships, \ie, temporal consistency. 
Since we want the edited frames (produced by CCS) to emulate the temporal consistency of the source frames, we constrain the edited frames to approximate the distribution of the source frames. 
This process can essentially be computed via Bayesian inference~\cite{box2011bayesian}. Given the complexity of video data, this inference occurs in a high-dimensional space, where the curse of dimensionality presents a long-standing challenge~\cite{wang2022projected}. To overcome it, we get inspiration from~\cite{kim2023collaborative}, and employ Stein Variational Gradient Descent (SVGD)~\cite{liu2016stein} to enhance the updating process, as SVGD offers the steepest descent for the updating, transforming the complex high-dimensional Bayesian inference into a deterministic updating process.
Below, we provide details on the SVGD-enhanced updating (TCS).

\noindent \textbf{Updating progress based on SVGD.} We consider the $N$ source frames as $N$ samples, $\{z(i)\}_{i=1}^N$, drawn from a distribution. We note that, as our method is based on the latent diffusion model, these samples $\{z(i)\}_{i=1}^N$  are in fact latents sampled from the latent space. We aim to update the CCS-generated samples $\{\hat{z}_0^{(0)}(i)\}_{i=1}^N$ (for simplicity, below we denote $\{\hat{z}_0^{(0)}(i)\}_{i=1}^N$ as $\{\hat{z}^{(0)}(i)\}_{i=1}^N$) based on SVGD to approximate the samples $\{z(i)\}_{i=1}^N$ of source frames, 
thereby resulting in edited frames with good temporal consistency. We refer to this updating process as TCS, which is performed after CCS is completed.
\input{tab/main_results}
Specifically, our TCS, based on SVGD, follows a deterministic process consisting of $\mathcal{L}$ recursive steps. Beginning from $ \ell = \mathcal{L} $, the $ \ell $-th step of the TCS denoises the latent representation $ \hat{z}^{(0)}_\ell(i) $ of the $ i $-th sample ($ i \in [1, \dots, N] $) into the latent $ \hat{z}^{(0)}_{\ell-1}(i) $ of the $ i $-th sample at the $ (\ell-1) $-th step:
\vspace{-1.5mm}
\begin{equation}
\begin{aligned}
\label{eq:updating process of SVGD}
&\hat{z}^{(0)}_{\ell-1}\left(i\right) = \hat{z}^{(0)}_\ell\left(i\right) - \eta \cdot \hat{\phi}\big(\hat{z}^{(0)}_\ell(i)\big), 
\text{where}~ \hat{\phi}\left(z\right) =  \frac{1}{N} \sum_{j=1}^{N} \big[ \\ &K\big(\hat{z}^{(0)}_\ell(j), z\big)    \big(\hat{z}^{(0)}_\ell(j)-z(j)\big)  + \nabla_{\hat{z}^{(0)}_\ell(j)} K\big(\hat{z}^{(0)}_\ell(j), z\big) \big].
\end{aligned}
\end{equation}
Here, $\eta$ is the step size, and $K(\cdot,\cdot)$ is a standard Radial Basis Function (RBF) kernel.  

In high-dimensional space, updating each sample based solely on its own gradient may lead to unstable optimization and increase the risk of getting trapped in local optima~\cite{ge2015escaping}. Therefore, we follow SVGD to use an averaged gradient across all $ N $ samples (represented by the first term in $\hat{\phi}\left(z\right)$) to update each sample. 
The second term in $ \hat{\phi}\left(z\right) $ serves as a repulsive force to prevent mode collapse among the samples, further contributing to the stability of the update process~\cite{liu2016stein}. The algorithm for our method is provided in \emph{supplementary}.

%% file: image/tex/G_i_.tex
\begin{figure}[t!]
  \centering
\includegraphics[width=0.48\textwidth]{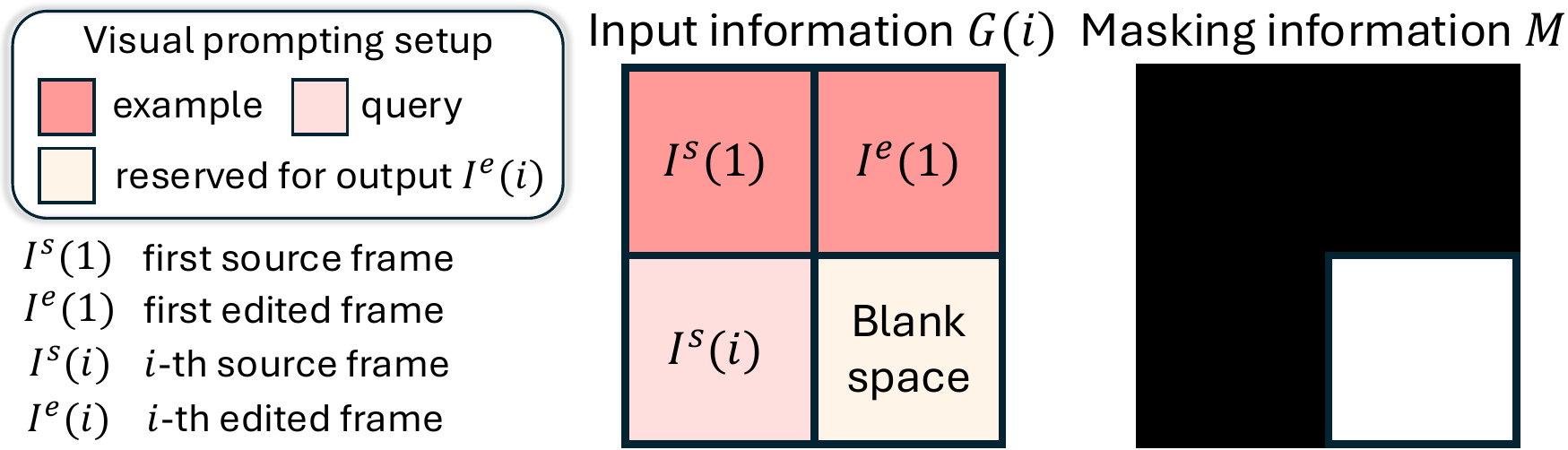}
  \caption{A visualization of  the input information $G(i)$ ($i$ denotes the $i$-th frame) and its corresponding mask information $M$.  
  }
  \label{fig:G(i)}
\end{figure}

%% file: tab/main_results.tex
\begin{table*}[t!]
\caption{Quantitative comparisons of our method with baselines.  \colorbox{tablered}{Best}  results are highlighted. ``$+$'' indicates that the metric is used to evaluate the edited region, whereas ``$-$'' indicates that the metric is used to evaluate the unedited region.  Following  Videoshop~\cite{fan2024videoshop}, we use Cotracker~\cite{karaev2023cotracker} to identify the edited and unedited regions. (T.C. = Temporal Consistency; E. = Efficiency)
}
\label{tab:main_results}
\centering
\resizebox{0.75\textwidth}{!}{
\begin{tabular}{c|ccc|cccccc|c|c}
        \toprule[2pt]
        \multirow{3}{*}{Method} & \multicolumn{3}{c|}{Edit Fidelity} & \multicolumn{6}{c|}{Source Faithfulness} & T.C. & E. \\

        & $\text{CLIP}_{\text{tar}} \uparrow$ 
        & $\text{CLIP}_{\text{tar}}^+ \uparrow$ 
        & $\text{TIFA} \uparrow$ 
        & $\text{CLIP}_{\text{src}} \uparrow$ 
        & $\text{CLIP}_{\text{src}}^+ \uparrow$ 
        & $\text{Flow}\downarrow$ 
        & $\text{Flow}^-\downarrow$ 
        & $\text{FVD}\downarrow$ 
        & $\text{SSIM}\uparrow$ 
        & $\text{CLIP}_{\text{TC}}\uparrow$  
        & time (s) $\downarrow$
        \\

        & $\left(\times 10^{-2}\right)$ 
        & $\left(\times 10^{-2}\right)$ 
        & $\left(\times 10^{-2}\right)$ 
        & $\left(\times 10^{-2}\right)$ 
        & $\left(\times 10^{-2}\right)$ 
        & $\left(\times 10^{-1}\right)$ 
        & $\left(\times 10^{-1}\right)$ 
        & $\left(\times 10^{2}\right)$ 
        & $\left(\times 10^{-2}\right)$ 
        & $\left(\times 10^{-2}\right)$ 
        & $\left(\times 10^{0}\right)$
        \\

        \hline
        BDIA~\cite{guoqiang2024bdia} & 82.1 & 82.2 & 57.7 & 82.5 & 87.1 & 28.3 & 14.3 & 34.8 & 49.7 & 94.4 & 35 \\
        Pix2Video~\cite{ceylan2023pix2video} & 71.2 & 76.5 & 52.0 & 74.6 & 79.0 & 35.9 & 25.8 & 29.9 & 59.1 & 94.5 & 157\\
        Fatezero~\cite{qi2023fatezero} & 84.9 & 79.1 & 55.4 & 92.4 & 86.9 & 44.2 & 31.1 & 22.1 & 48.6 & 95.7 & 25 \\
        Spacetime~\cite{yatim2024space} & 63.9 & 75.2 & 46.3 & 65.7 & 71.9 & 82.4 & 56.2 & 48.2 & 41.6 & 96.6 & 135\\
        RAVE~\cite{kara2024rave} & 74.7 & 78.6 & 51.1 & 76.0 & 80.2 &33.5 & 24.2 & 23.5 & 62.2 & 96.6 & 45\\
        AnyV2V~\cite{blattmann2023stable} & 87.1 & 85.9 & 67.0 & 91.3 & 94.2 & 24.6 & 14.1 & 17.1 & 65.5 & 93.9 & 149\\
        Videoshop~\cite{fan2024videoshop} & 88.8 & 85.6 & 64.4 & 91.0 & 94.8 & \cellcolor{tablered}19.0 & \cellcolor{tablered}7.8 & \cellcolor{tablered}14.8 & \cellcolor{tablered}71.9 & 95.2 & 32\\
        \hline

        Ours w/o CCS & 80.3 & 77.6 & 55.8 & 81.3 & 82.7 & 23.7 & 10.9 & 25.1 & 59.8 & 95.8 & 19\\ 
        
        Ours w/o TCS & 89.8 & 86.1 & 68.3 & 92.8 & 95.1 & 33.1 & 20.5 & 23.7 & 69.5 & 89.8 & \cellcolor{tablered}18\\
                Ours & \cellcolor{tablered}{90.1} & \cellcolor{tablered}88.2 & \cellcolor{tablered}69.1 & \cellcolor{tablered}93.2 & \cellcolor{tablered}96.6 & 21.9 & 9.2 & 15.2 & 69.2 & \cellcolor{tablered}97.1 & 19\\   
        \toprule[2pt] 
          
\end{tabular}
}
\end{table*}

%% file: real_sec/5_experiment.tex
\section{Experiments}
\label{subsec:Experiments}

\subsection{Experiment setup}

\noindent \textbf{Implementation details.} We use Stable Diffusion Inpainting 1.5~\cite{inpaint2022} as the image inpainting diffusion model. CCS operates with 30 timesteps, and TCS utilizes 50 timesteps.  We set $\lambda_1 = 0.7$, $\lambda_2 = 1.2$, and $\eta = 2.0$. We follow~\cite{gu2024analogist} to use the self-attention cloning in our CCS.
All experiments are conducted on a A100 GPU.

\input{image/tex/com}

\noindent \textbf{Datasets.}  Following Videoshop~\cite{fan2024videoshop}, our method is evaluted  on a large-scale generated video dataset derived from MagicBrush dataset~\cite{zhang2024magicbrush}, which consists of 10388 tuples in the format \emph{(source video, editing instruction, the first edited frame)}. These tuples represent a wide range of editing types, including object addition, replacement, removal, and modifications in action, color, and texture, \etc.  

\noindent \textbf{Baselines.} Our method is compared with 2 state-of-the-art (SOTA) OCVE methods (Videoshop~\cite{fan2024videoshop}, AnyV2V~\cite{ku2024anyv2v}). Following  Videoshop, we also compare our method with 5 SOTA text-based video editing methods: Pix2Video~\cite{ceylan2023pix2video}, Fatezero~\cite{qi2023fatezero}, Spacetime~\cite{yatim2024space}, RAVE~\cite{kara2024rave}, and BDIA~\cite{guoqiang2024bdia}.

\noindent \textbf{Evaluation metrics.}  Following Videoshop~\cite{fan2024videoshop}, we  evaluate our method from 4 perspectives. \textbf{1)} Edit fidelity: We measure $\text{CLIP}_{\text{tgt}}$ similarity~\cite{radford2021learning}  between each edited frame and the first edited image. We use the TIFA score~\cite{hu2023tifa} to assess the semantic alignment between the first edited frame and subsequent edited frames in the video. \textbf{2)} Source faithfulness: We measure $\text{CLIP}_{\text{src}}$ similarity between the source and edited videos. Flow score~\cite{teed2020raft} is employed to evaluate motion faithfulness. The FVD and SSIM scores are used to assess the overall quality of the edited videos and the quality of edited frames, respectively. \textbf{3)} Temporal consistency: We measure the average CLIP similarity between adjacent frames, referred to as $\text{CLIP}_{\text{TC}}$. \textbf{4)}  Efficiency: We measure the average time taken by each video editing method to process a video.  \textbf{5)} Human evaluation: We ask human evaluators to compare the editing quality of our method with that of the baseline. For more details of the metrics, please refer to our \textit{supplementary}.

\subsection{Quantitative results}
We compare our method with SOTA video editing methods~\cite{ku2024anyv2v,fan2024videoshop} on the generated dataset, as shown in~\cref{tab:main_results}. Our proposed method achieves the best performance across multiple  metrics, demonstrating its effectiveness. The time metric clearly shows that our method is significantly more efficient compared to previous OCVE methods~\cite{ku2024anyv2v,fan2024videoshop}. This efficiency arises from our use of a more streamlined image diffusion model, rather than the video diffusion models used in earlier OCVE approaches. Furthermore, SVGD used in our CCS method, which ensures temporal consistency across the edited frames, is also efficient~\cite{liu2016stein}. We provide results of the human evaluation in \emph{supplementary}.

\subsection{Qualitative results}
As shown in~\cref{fig:com}, we evaluate our method with 2 SOTA OCVE methods~\cite{ku2024anyv2v,fan2024videoshop} on two types of edits: object replacement and object removal/addition. Our method achieves the best performance in both cases. For instance, in the edit where fruit is replaced with vegetables, Videoshop struggles to maintain the  fruit's appearance. In contrast, our method, leveraging CCS, consistently preserves the appearance of the fruit across the edited frames.

\subsection{Ablation studies}
We evaluates two variants of our method: ``Ours w/o CCS'' and ``Ours w/o TCS'' (see~\cref{tab:main_results}).  By comparing the source faithfulness metrics between ``Ours w/o CCS'' and ``Ours'', we observe a performance drop in ``Ours w/o CCS'', indicating that our designed CCS effectively enhances content consistency between the edited frames and the source frames.  The temporal consistency metric for ``Ours w/o TCS'' shows a significant decrease, demonstrating that TCS effectively improves temporal consistency across the edited frames.

%% file: image/tex/com.tex
\begin{figure*}[t!]
  \centering
\includegraphics[width=1\textwidth]{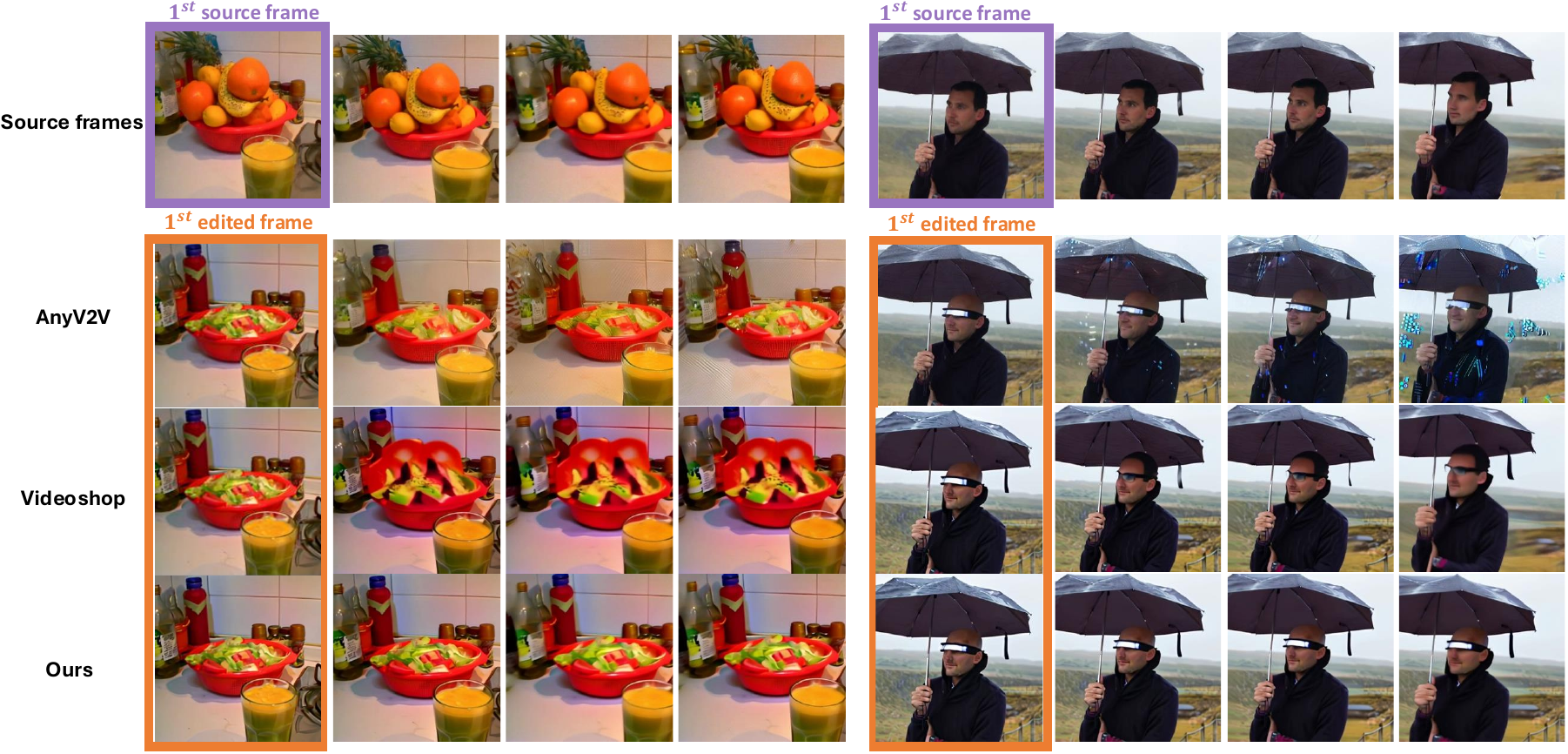}
  \caption{The visual comparison includes our method alongside two SOTA OCVE methods (AnyV2V~\cite{ku2024anyv2v} and Videoshop~\cite{fan2024videoshop}), evaluated across two distinct types of editing. On the left, user modifications consist of replacing the fruit in the basin with vegetables. On the right, the user edits involve: 1) removing the individual's hair and 2) adding glasses.}
  \label{fig:com}
\end{figure*}

%% file: real_sec/6_conclusion.tex
\section{Conclusion}
In this paper, we perform OCVE via visual prompting, eschewing the  DDIM inversion process  which can potentially introduce errors. Our method comprises CCS and TCS, which ensure content consistency between the edited and source frames, as well as maintain temporal consistency throughout the edited frames. Both quantitative and qualitative experimental results validate the efficacy of our method.